\documentclass[letterpaper, 10 pt, journal, twoside]{IEEEtran}

\usepackage{graphicx, tipa}
\usepackage{amsmath,amssymb,amsfonts}
\usepackage[caption=false]{subfig}
\usepackage[ruled,linesnumbered, noend]{algorithm2e}
\usepackage{dirtytalk}
\usepackage{hyperref}
\usepackage{gensymb}
\usepackage{tabularx}
\usepackage{multirow}
\usepackage[short]{optidef}
\usepackage{booktabs}
\usepackage{siunitx}
\usepackage{titlesec}
\newcounter{cases}
\newcounter{subcases}[cases]

\makeatletter
\newcommand{\removelatexerror}{\let\@latex@error\@gobble}
\makeatother
\SetKwComment{Comment}{$\triangleright$\ }{}

\newcommand\Tstrut{\rule{0pt}{2.0ex}}         

\title{NavRL: Learning Safe Flight in Dynamic Environments}

\author{Zhefan Xu, Xinming Han, Haoyu Shen, Hanyu Jin, and Kenji Shimada%
\thanks{Manuscript received: September, 23, 2024; Revised January, 3, 2025; Accepted February, 11, 2025.}
\thanks{This paper was recommended for publication by Editor Aleksandra Faust upon evaluation of the Associate Editor and Reviewers' comments.} 
\thanks{Zhefan Xu, Xinming Han, Haoyu Shen, Hanyu Jin and Kenji Shimada are with the Department of Mechanical Engineering, Carnegie Mellon University, 5000 Forbes Ave, Pittsburgh, PA, 15213, USA. {\tt\footnotesize zhefanx@andrew.cmu.edu}}%
\thanks{Digital Object Identifier (DOI): see top of this page.}
}

\begin{document}
\markboth{IEEE Robotics and Automation Letters. Preprint Version. Accepted February, 2025}
{Xu \MakeLowercase{\textit{et al.}}: NavRL: Learning Safe Flight in Dynamic Environments} 

\maketitle

\noindent \begin{abstract}
Safe flight in dynamic environments requires unmanned aerial vehicles (UAVs) to make effective decisions when navigating cluttered spaces with moving obstacles. Traditional approaches often decompose decision-making into hierarchical modules for prediction and planning. Although these handcrafted systems can perform well in specific settings, they might fail if environmental conditions change and often require careful parameter tuning. Additionally, their solutions could be suboptimal due to the use of inaccurate mathematical model assumptions and simplifications aimed at achieving computational efficiency. To overcome these limitations, this paper introduces the NavRL framework, a deep reinforcement learning-based navigation method built on the Proximal Policy Optimization (PPO) algorithm. NavRL utilizes our carefully designed state and action representations, allowing the learned policy to make safe decisions in the presence of both static and dynamic obstacles, with zero-shot transfer from simulation to real-world flight. Furthermore, the proposed method adopts a simple but effective safety shield for the trained policy, inspired by the concept of velocity obstacles, to mitigate potential failures associated with the black-box nature of neural networks. To accelerate the convergence, we implement the training pipeline using NVIDIA Isaac Sim, enabling parallel training with thousands of quadcopters. Simulation and physical experiments\footnote{Experiment video link: \url{https://youtu.be/EbeJW8-YlvI}} show that our method ensures safe navigation in dynamic environments and results in the fewest collisions compared to benchmarks.
\end{abstract}

\section{Introduction}

\IEEEPARstart{A}{utonomous} unmanned aerial vehicles (UAVs) are widely used in applications like exploration \cite{zhou2021fuel}, search and rescue \cite{alsamhi2022uav}, and inspection \cite{inspection}. These tasks often occur in dynamic environments that require effective collision avoidance. Traditional methods rely on handcrafted algorithms and hierarchical modules, leading to overly complex systems with hard-to-tune parameters. In contrast, reinforcement learning (RL) allows UAVs to learn decision-making through experience, offering better adaptability and improved performance. Developing RL-based navigation methods is essential for enhancing UAV safety in dynamic environments.

\begin{figure}[t] 
    \vspace{0.1cm}
    \centering
    \includegraphics[scale=0.52]{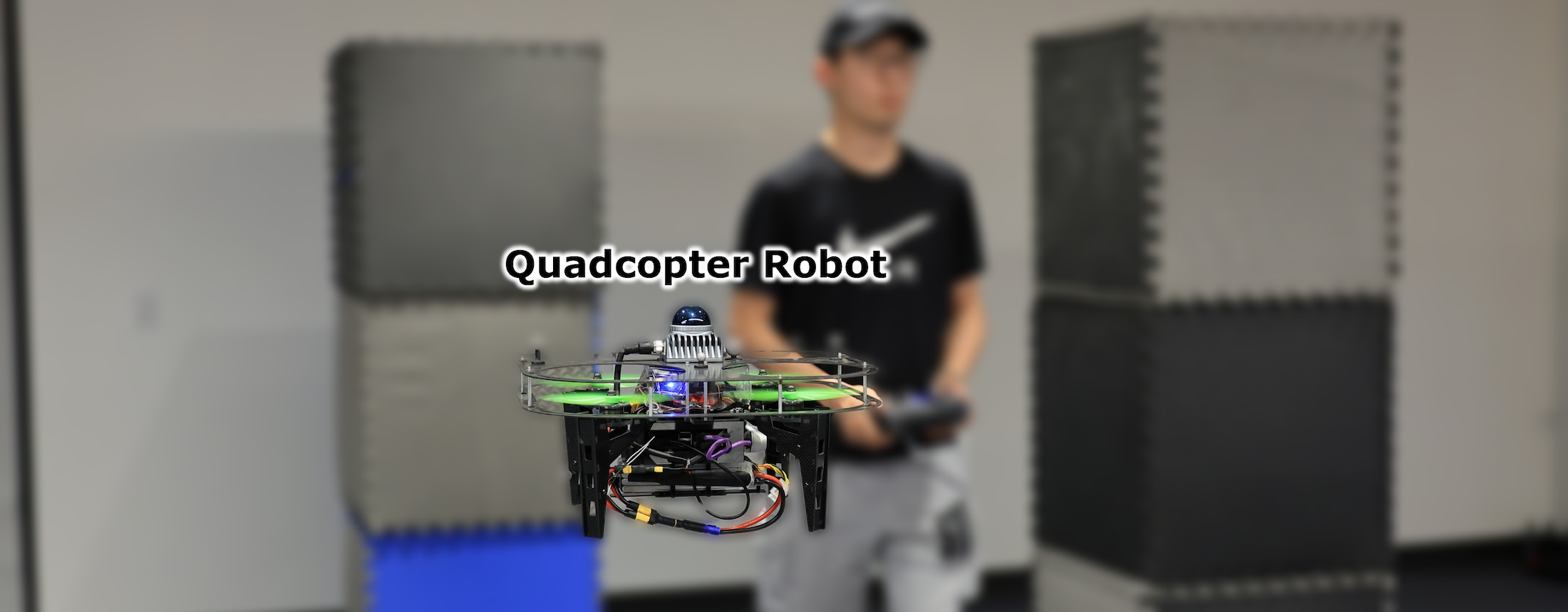}
    \caption{A customized quadcopter UAV navigating a dynamic environment using the proposed NavRL framework. The robot achieves safe navigation and effective collision avoidance with both static and dynamic obstacles.}
    \label{intro-figure}
\end{figure}

Developing an RL-based navigation method suitable for real-world deployment presents several challenges. First, since reinforcement learning involves training the robot through collision experiences, the learning process must occur in simulated environments. This creates a sim-to-real transfer issue due to the gap between simulated and real-world sensory information, particularly with camera images. Previous works have attempted to address this issue by developing methods to reduce the gap between simulated and real-world environments \cite{kulkarni2023task}\cite{hoeller2021learning}\cite{song2023learning} or by training the robot directly in real-world settings \cite{gandhi2017learning}\cite{smolyanskiy2017toward}\cite{tolani2021visual}. However, these approaches often require additional training steps and can be data-inefficient when trained in real-world scenarios. Second, even if the trained RL policy demonstrates satisfactory performance, ensuring safety remains challenging due to the black-box nature of neural networks, necessitating an effective safety mechanism to prevent severe failures \cite{ray2019benchmarking}\cite{thananjeyan2021recovery} \cite{hsu2023sim}\cite{kochdumper2023provably}. Lastly, training a reinforcement learning policy requires a substantial amount of robot experience, and some previous methods \cite{sadeghi2016cad2rl}\cite{xie2017towards}\cite{brilli2023monocular} that collect data using a single robot often result in slow convergence speeds due to limited data diversity and reduced parallel exploration opportunities.

To address these issues, this paper proposes a deep reinforcement learning-based navigation framework, named NavRL, based on the Proximal Policy  Optimization (PPO) algorithm \cite{schulman2017proximal}. The proposed framework employs state and action representations built on our perception module, specifically designed for collision avoidance in dynamic environments, enabling zero-shot sim-to-real transfer capability. Additionally, to prevent severe failures, we leverage the concept of velocity obstacles (VO) \cite{velocity_obstacles} to create a simple but effective safety shield using linear programming to optimize the RL policy network's action outputs. To accelerate training convergence, we design a parallel training pipeline capable of simulating thousands of quadcopters simultaneously using NVIDIA Isaac Sim. We validate the proposed framework through extensive simulation and physical flight experiments in various environments, demonstrating its ability to ensure safe navigation. Fig. \ref{intro-figure} illustrates an example of our UAV navigating in a dynamic environment using the proposed NavRL framework. The main contributions of this work are:
\begin{itemize}
    \item \textbf{The NavRL Navigation Framework:} This work introduces a novel reinforcement learning-based UAV navigation system to ensure safe autonomous flight in dynamic environments. The NavRL navigation framework is made available as an open-source package on GitHub \footnote{Software available at: \url{https://github.com/Zhefan-Xu/NavRL}}.
    \item \textbf{Policy Action Safety Shield:} Our method adopts a safety shield into the policy network's action outputs to enhance safety based on the velocity obstacle concept.
    \item \textbf{Physical Flight Experiments:} Real-world experiments in various environments are conducted to demonstrate the safe navigation capabilities and zero-shot sim-to-real transfer effectiveness of the proposed method. 
\end{itemize}

\section{Related Work}

Research on UAV navigation in dynamic environments often relies on rule-based methods with handcrafted algorithms \cite{zju_collision_avoidance}\cite{ViGO}\cite{risk-sampling}, which can be complex and require careful tuning as conditions change. In contrast, learning-based methods reduce complexity and adapt better to varying environments. This section mainly categorizes learning-based navigation into supervised and reinforcement learning-based methods, while acknowledging the existence of other approaches.

\textbf{Supervised learning-based methods:} Methods in this category train networks using labeled datasets. Early approaches \cite{smolyanskiy2017toward}\cite{tai2016deep} deploy robots in real-world environments, collect images, and manually label them with ground truth actions. Similarly, some methods \cite{gandhi2017learning}\cite{padhy2018deep} predict the safety of input images rather than outputting the decision and then use handcrafted algorithms to control the robot. 

The methods mentioned above require manually labeled real-world data and often suffer from limited generalization ability due to insufficient data. Loquercio et al. \cite{loquercio2018dronet} train the network on an autonomous driving dataset to achieve navigation, benefiting from its extensive data volume. Jung et al. \cite{jung2018perception} use a learning-based detector to enable a drone to pass through gates in racing, while in \cite{lv2023autonomous}, iterative learning control is applied to reduce tracking error in this context. Works on foundation models for visual navigation has been developed using real-world experience from various robots \cite{shah2023vint}\cite{shah2023gnm}\cite{sridhar2024nomad}. Simon et al. \cite{simon2023mononav} demonstrate collision avoidance capabilities for drones equipped with monocular cameras using a depth estimation method \cite{bhat2023zoedepth}. An imperative learning approach based on semantic images is proposed in \cite{roth2024viplanner} to achieve semantically-aware local navigation.

\textbf{Reinforcement learning-based methods:} Compared to supervised learning-based methods, reinforcement learning-based approaches benefit from the abundant data in simulation. Some methods utilizing Q-learning \cite{sadeghi2016cad2rl}\cite{xie2017towards}\cite{singla2019memory} or value learning \cite{chen2017decentralized}\cite{chen2017socially} have demonstrated successful navigation. However, these methods are constrained to discrete action spaces, which can lead to suboptimal performance. 

Policy gradient methods \cite{brilli2023monocular}\cite{everett2018motion}, using an actor-critic structure, have gained popularity for enabling robot control in continuous action spaces. Kaufmann et al. \cite{kaufmann2023champion} trained a policy in simulation to outperform human champions in drone racing, with other studies enhancing control performance \cite{bauersfeld2023user}\cite{romero2024actor}. Song et al. \cite{song2023learning} developed a vision-based network by distilling knowledge from a teacher policy with privileged information, while Xing et al. \cite{10610095} applied contrastive learning to improve image encoding. In \cite{OmniDrones}, a comprehensive reinforcement training pipeline is presented for UAV-related tasks.


To ensure safety, a recovery policy using a reach-avoid network is introduced in \cite{AgileButSafe} to prevent failures. Similarly, Kochdumper et al. \cite{kochdumper2023provably} utilize reachability analysis to project unsafe policy actions to safe regions. However, their approach requires a precomputed reachability set, with computation costs scaling exponentially with action dimensions.

Most existing methods are designed for navigation and collision avoidance in static environments. Some methods, such as \cite{everett2018motion}, demonstrate dynamic obstacle avoidance but do not address collision avoidance in complex static scenarios. The method in \cite{AgileButSafe} can handle both static and dynamic obstacles but may perform suboptimally by treating dynamic obstacles as static. Recent safety-focused approaches are either computationally expensive or require extensive training. These challenges, combined with the difficulty of sim-to-real transfer, motivate us to propose a framework that ensures safe navigation while avoiding both static and dynamic obstacles.

\section{Methodology}
\begin{figure*}[t] 
    \centering
    \includegraphics[scale=0.78]{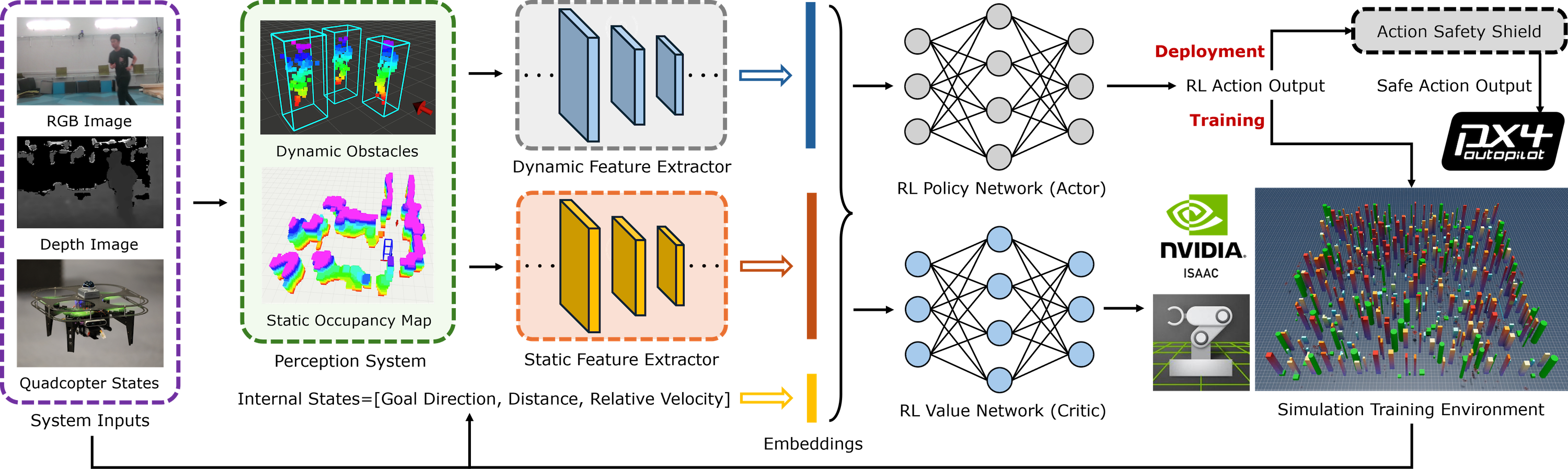}
    \caption{The proposed NavRL framework. The perception system processes RGB-D images along with the robot's internal states to generate representations for both static and dynamic obstacles. These representations are then fed into two feature extractors, which produce state embeddings concatenated with the robot's internal states. In the training phase, an actor-critic network structure is utilized to train robots in parallel within the NVIDIA Isaac Sim environment. During the deployment stage, the policy network generates actions that are further refined by a safety shield mechanism to ensure safe robot control. }
    \label{system-overview}
\end{figure*}

The proposed NavRL navigation framework is depicted in Fig. \ref{system-overview}. The obstacle perception system processes RGB-D images and robot states to generate representations for both static and dynamic obstacles (Sec. \ref{perception system}). Section \ref{RL formulation} details the conversion of these obstacle representations into network input states, along with the definitions of robot actions and training rewards. During training, we employ the PPO algorithm with an actor-critic network structure to train the policy and value networks, as explained in Sec. \ref{network and training}. For deployment, a safety shield is applied to the RL policy network's action outputs to ensure safety (Sec. \ref{safety shield}).

\subsection{Obstacle Perception System} \label{perception system}
Static and dynamic obstacles are handled separately by our perception system due to their different properties, as illustrated in Fig. \ref{system-overview}. Static obstacles, which can have arbitrary shapes and sizes, are represented more accurately using a discretized format, such as an occupancy voxel map. Conversely, dynamic obstacles, typically modeled as rigid bodies, are represented using bounding boxes with estimated velocity information. The rationale for using different representations for static and dynamic obstacles lies in their distinct characteristics. Static obstacles sometimes have irregular geometries, making a discrete representation more suitable for accurate shape description. In contrast, dynamic obstacles, typically rigid bodies, require a representation that captures their states as a whole entity. Additionally, updating the occupancy map with dynamic obstacles can introduce noise and latency, further justifying the need for separate representations. For the perception of static obstacles, we create a 3D occupancy voxel map with a fixed memory size, determined by the maximum number of voxels suitable for the environment. The occupancy data for each voxel is then stored in a pre-allocated array. This design enables us to access occupancy information with constant time complexity ($\mathcal{O}(1)$). At each time step, we recursively update the log probability of occupancy for each voxel based on the latest depth image and clear the occupancy data for dynamic obstacles by iterating through their detected bounding boxes. It is important to note that this static occupancy map can be generated on the fly without the need for any prebuilt data.

Detecting 3D dynamic obstacles is challenging due to the noisy depth images from the lightweight UAV camera and the limited processing power of the onboard computer. To achieve accurate detection with minimal computational demand, we propose an ensemble method that combines two lightweight detectors built on \cite{detector}. The first, the U-depth detector \cite{early_reactive}, converts raw depth images into a U-depth map (similar to a top-down view) and uses a contiguous line grouping algorithm to detect 3D bounding boxes of obstacles. The second, named the DBSCAN detector, applies the DBSCAN clustering algorithm to point cloud data from depth images to identify obstacle centers and dimensions by analyzing boundary points within each cluster. Both detectors, however, can produce a significant number of false positives due to the noisy input data. Our proposed ensemble method addresses this by finding the mutual agreements from both detectors to identify consistent results. Furthermore, to differentiate between static and dynamic obstacles, we employ a lightweight YOLO detector to classify dynamic obstacles by examining the 2D bounding boxes that are re-projected onto the image plane from the 3D detections. 

The dynamic obstacle velocity is estimated by the tracking module, which operates in two stages: data association and state estimation. In the data association stage, the goal is to establish correspondences between detected obstacles across consecutive time steps. To minimize detection mismatches, we construct a feature vector for each obstacle that includes its position, bounding box dimensions, point cloud size, and point cloud standard deviation, and then determine matches based on the similarity scores between potential matching obstacles. In the state estimation stage, a Kalman filter is used to estimate obstacle velocities, with a constant acceleration model applied to account for changes in velocity over time. It is worth noting that while the perception system in this framework is based on a camera, other sensors can also be used if they provide a similar obstacle representation. The dynamic obstacle detection and tracking in the perception system is based on the approach presented in \cite{detector}, where readers can find further details and implementation specifics.

\subsection{Reinforcement Learning Formulation} \label{RL formulation}
The navigation task is formulated as a Markov Decision Process (MDP) defined by the tuple $(S, A, P, R, \gamma)$, where $S$ is the state space (robot's internal and sensory data), $A$ is the action space, the transition function $P(s_{t+1}|s_{t}, a_{t})$ models environment dynamics, and the reward function $R(s_{t}, a_{t})$ encourages goal-reaching behaviors while penalizing collisions and inefficient actions. The goal is to learn an optimal policy $\pi^{*}(a_{t}|s_{t})$ that maximizes the expected cumulative reward:

\begin{equation}
    \pi^* = \arg \max_{\pi} \mathbb{E} \left[ \sum_{t=0}^{T} \gamma^t R(s_t, a_t) \right],
\end{equation}
where $\gamma \in [0, 1]$ is the discount factor for future rewards. This formulation enables the use of reinforcement learning to train the robot for safe navigation in dynamic environments.


\textbf{State:} As the input to the policy, the state must include all information relevant to navigation and collision avoidance. In our system framework shown in Fig. \ref{system-overview}, the designed state is composed of three parts: the robot's internal states, dynamic obstacles, and static obstacles. The robot's internal states provide details about the robot's direction and distance to the navigation goal with its current velocity defined as:
\begin{equation}
    S_{int} = [\frac{P^{G}_{g} - P^{G}_{r}}{\lVert P^{G}_{g} - P^{G}_{r} \rVert}, \lVert P^{G}_{g} - P^{G}_{r} \rVert, V^{G}_{r}]^T,
\end{equation}
where $P_{r}$ and $P_{g}$ represent the robot position and goal position, respectively, and $V_{r}$ is the robot current velocity. The superscript $(\cdot)^{G}$ indicates that the vector is expressed in the \say{goal coordinate frame}, which is defined with its origin at the robot's starting position. In this frame, the x-axis aligns with the vector pointing from the starting position, $P_{s}$, to the goal position, while the y-axis lies parallel to the ground plane.  This goal coordinate transformation reduces dependency on the global coordinate system, improving overall RL training convergence speed, and will also be applied to the definition of obstacle state representations. 

The dynamic obstacles are represented using a 2D matrix:
\begin{equation}
    S_{dyn} = [\mathcal{D}_{1}, ..., \mathcal{D}_{N_{d}}]^{T}, \ S_{dyn} \in \mathbb{R}^{N_{d} \times M}, \ \mathcal{D}_{i} \in \mathbb{R}^{M}.
\end{equation}
In this formulation, $\mathcal{D}_{i}$ denotes the state vector of the $i$th closest dynamic obstacle to the robot and is expressed as:
\begin{equation}
    \mathcal{D}_{i} = [\frac{P^{G}_{o_{i}} - P^{G}_{r}}{\lVert P^{G}_{o_{i}} - P^{G}_{r} \rVert}, \lVert P^{G}_{o_{i}} - P^{G}_{r} \rVert, V^{G}_{o_{i}}, \text{dim}({o_{i}}) ]^{T},
\end{equation}
where $P_{o_{i}}$ and $V_{o_{i}}$ represent the center position and velocity of the dynamic obstacle, respectively, and $\text{dim}(o_{i})$ indicates the height and width of the obstacle. The number of dynamic obstacles, $N_{d}$, is predefined, and if the actual number of detected obstacles is less than this predefined limit, the state vector values are set to zero.  The relative position vectors for both internal states and obstacle states are split into a unit vector and its norm, as this method showed slightly faster convergence speed in our experiments.

\begin{figure}[t] 
    \centering
    \includegraphics[scale=0.40]{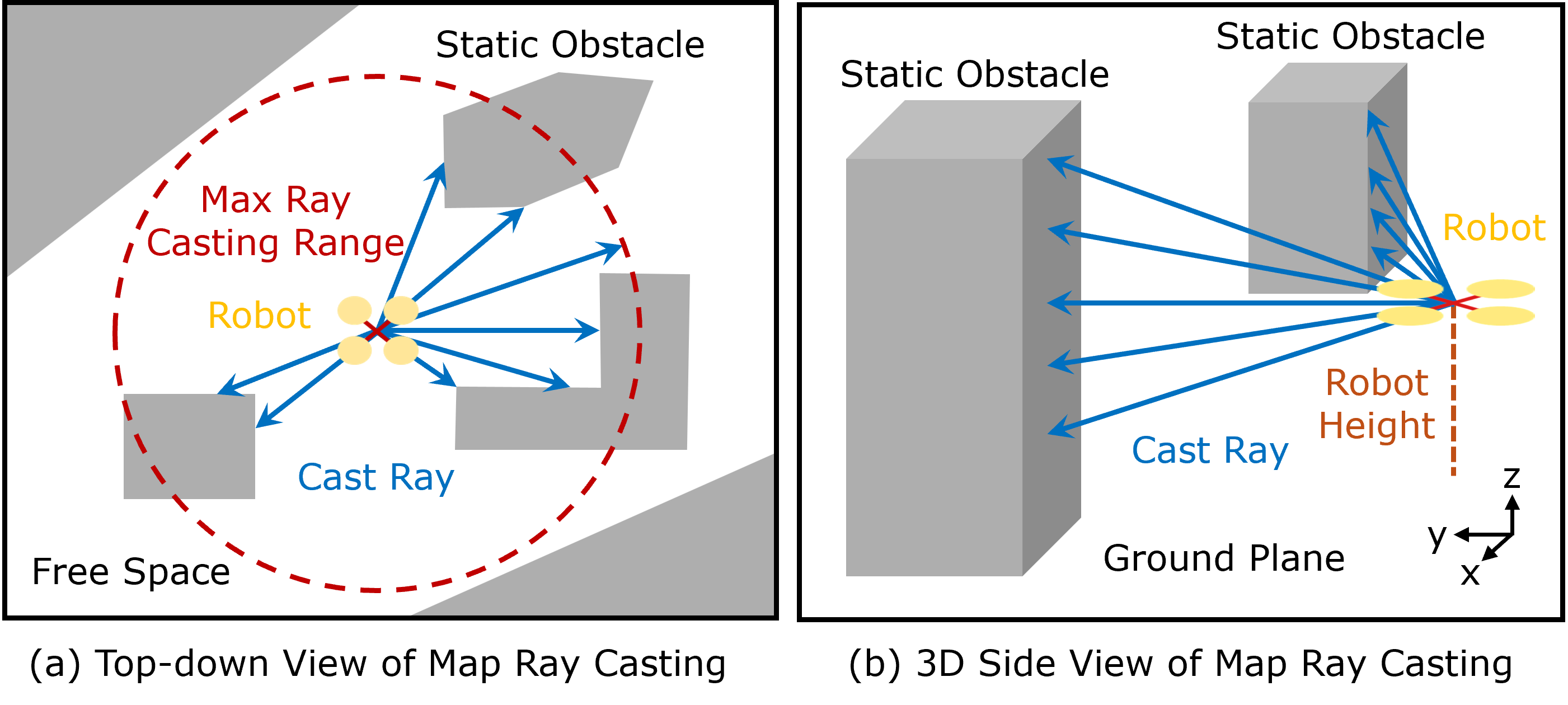}
    \caption{Illustration of map ray casting. Only rays within the maximum range are shown. (a) A top-down view of horizontally cast rays with a 360-degree casting angle. (b) A side view displaying rays in the vertical planes.} 
    \label{map-raycast}
\end{figure}

Compared to dynamic obstacles, static obstacles are represented as map voxels, which cannot be directly input into neural networks. Therefore, we perform 3D ray casting from the robot's position against the map.  In Fig. \ref{map-raycast}a, rays are cast horizontally in all directions within a maximum range using a user-defined ray casting angle interval. Similarly, Fig. \ref{map-raycast}b illustrates the same operation in the vertical plane. For each diagonal ray casting angle $\theta_{i}$ in the vertical plane, the lengths of all rays in the horizontal plane are recorded into a vector $R_{\theta_{i}}$. Any ray exceeding the maximum range is assigned a length equal to the maximum range plus a small offset, allowing obstacle absence to be identified. The representation of static obstacles is constructed by stacking the ray length vectors for all diagonal ray angles in the vertical planes:
\begin{equation}
    S_{stat} = [R_{\theta_{0}}, ..., R_{\theta_{N_{v}}}], S_{stat} \in \mathbb{R}^{N_{h} \times N_{v}}, R_{\theta_{i}} \in \mathbb{R}^{N_{h}},  \label{static obstacle states}
\end{equation}
where $N_{v}$ and $N_{h}$ represent the number of rays in the vertical and horizontal planes, determined by the ray casting angle interval and the vertical field of view. Unlike using image data as state representation, the proposed RL state representation has minimal discrepancies between simulation and the real world, which is beneficial for sim-to-real transfer.

\begin{figure}[t] 
    \centering
    \includegraphics[scale=0.087]{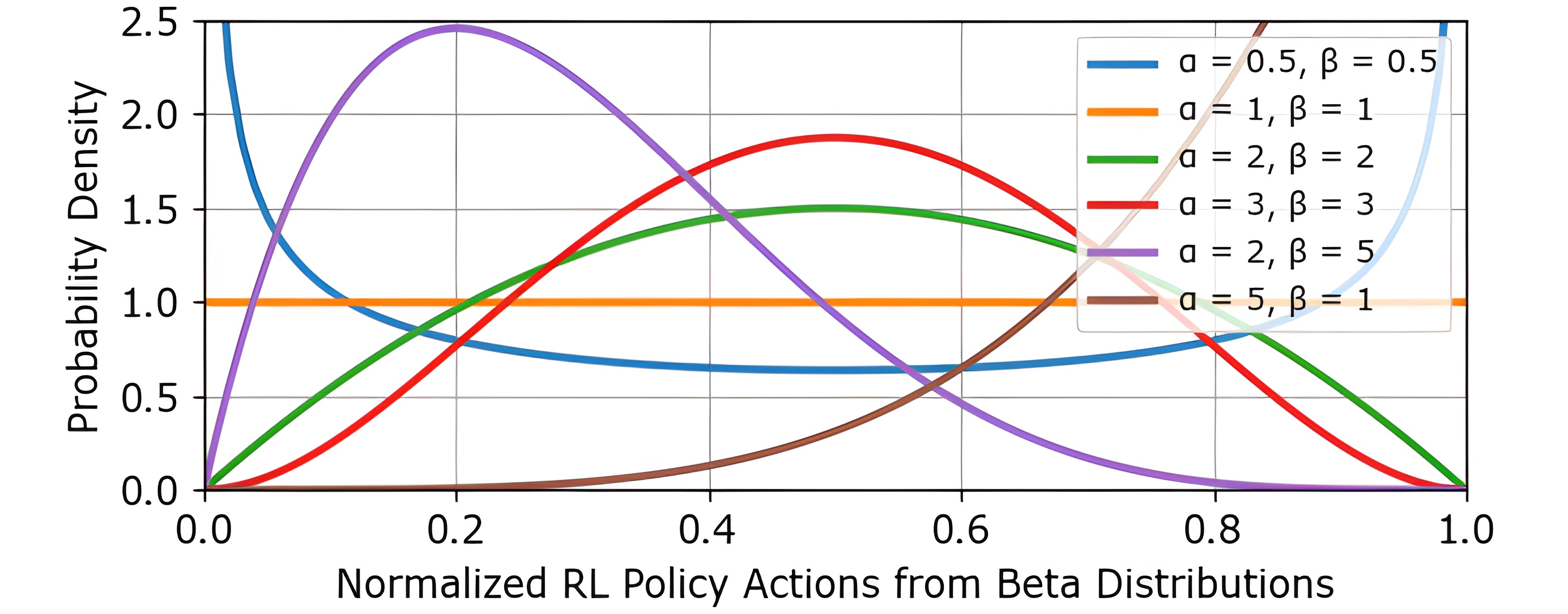}
    \caption{Visualization of example RL policy actions from Beta distributions.}
    \label{beta-distrbution}
\end{figure}

\textbf{Action:} At each time step, the velocity control $V_{ctrl} \in \mathbb{R}^{3}$ is provided to the robot for navigation and collision avoidance. Velocity control is chosen because higher-level controls offer better transferability and generalization across different platforms, facilitating sim-to-real transfer. Additionally, velocity commands are more interpretable and easier for humans to supervise compared to lower-level controls. Instead of directly outputting velocity values from the RL policy, the policy is designed to first infer a normalized velocity $\hat{V}^{G}_{ctrl}$ with the final output velocity expressed as:
\begin{equation}
    V^{G}_{ctrl} = v_{lim} \cdot (2 \cdot \hat{V}^{G}_{ctrl} - 1)  , \ \hat{V}^{G}_{ctrl} \in [0, 1],
\end{equation}
where $v_{lim}$ is the user-defined maximum velocity. The final velocity is expressed in the goal coordinate frame, as described in the state formulation, requiring a coordinate transformation to be applied to the robot. This approach offers greater flexibility compared to other formulations where the RL policy must learn the action limits and cannot easily adjust the trained action range. To constrain the network output to the range $[0, 1]$, the model is designed to produce parameters ($\alpha, \beta$) for a Beta distribution, as shown in Fig. \ref{beta-distrbution}. When the RL action space is constrained, a Beta distribution-based policy has been proven to be bias-free and achieve faster convergence compared to a Gaussian distribution-based policy \cite{chou2017improving}. During the training process, exploration is encouraged by sampling from the Beta distribution using the generated parameters. In deployment, we use the mean of the Beta distribution as the normalized velocity output.

\textbf{Reward:} The designed RL reward function is computed at each time step consisting of multiple components: 
\begin{equation}
    r = \lambda_{1} r_{vel} + \lambda_{2} r_{ss} + \lambda_{3} r_{ds}  + \lambda_{4} r_{smooth} + \lambda_{5} r_{height}, 
\end{equation}
where $r_{(\cdot)}$ represents one type of reward weighted by $\lambda_{i}$. Each reward will be explained in the following paragraphs.

\textit{a) velocity reward $r_{vel}$}: The velocity reward encourages the robot to adopt velocities leading to the goal position:
\begin{equation}
    r_{vel} = \frac{P_{g} - P_{r}}{\lVert P_{g} - P_{r} \rVert} \cdot V_{r}, \ \text{where} \  P_{g}, P_{s}, V_{r} \in \mathbb{R}^3.
\end{equation}
This formulation rewards velocities that align more closely with the position-to-goal direction with higher speeds.

\textit{b) static safety reward $r_{ss}$}: The static safety reward ensures that the robot keeps a safe distance from static obstacles. Given the static obstacle states in Eqn. \ref{static obstacle states}, it is defined as:
\begin{equation}
    r_{ss} = \frac{1}{N_{h}N_{v}} \sum^{N_h}_{i=1} \sum^{N_v}_{j=1} \log S_{stat}(i, j). 
\end{equation}
This formulation computes the average log distance to static obstacles using the ray distances. The reward is maximized when the robot maintains greater distances from obstacles.

\textit{c) dynamic safety reward $r_{ds}$}: Similar to the static safety reward, the dynamic safety reward encourages the robot to avoid dynamic obstacles and is expressed as:
\begin{equation}
    r_{ds} = \frac{1}{N_{d}} \sum^{N_{d}}_{i=1} \log \lVert P_{r} - P_{o_{i}} \rVert. 
\end{equation}

\textit{d) smoothness reward $r_{smooth}$}: The smoothness reward penalizes sudden changes in the control output, written as:
\begin{equation}
    r_{smooth} = -\lVert V_{r}(t_{i}) - V_{r}(t_{i-1}) \rVert,
\end{equation}
where the L2 norm of the difference between the robot's velocities at the current and previous time steps is computed.

\textit{e) height reward $r_{height}$}: The height reward is designed to prevent the robot from avoiding obstacles by flying excessively high. It can be written as the following:
\begin{equation}
    r_{height} = -(min(|P_{r, z} -  P_{s, z}|, |P_{r, z} -  P_{g, z}|))^2,
\end{equation}
which applies when the current height $P_{r, z}$ falls outside the range defined by the start height $P_{s, z}$ and goal height $P_{g, z}$.

\subsection{Network Design and Policy Training} \label{network and training}
Given that our state representation consists of multiple components, a preprocessing step is necessary before inputting the data into the RL policy network. Both static and dynamic obstacles are represented as 2D matrices, so we utilize convolutional neural networks (CNN) to extract their features and transform them into 1D feature embeddings, as illustrated in Fig. \ref{system-overview}. These embeddings are then concatenated with the robot's internal states to form the complete input feature for the policy and value networks. We employ the Proximal Policy Optimization (PPO) algorithm \cite{schulman2017proximal} to train the actor (policy) and critic (value) networks, both of which are implemented using multi-layer perceptron. The training process is conducted in NVIDIA Isaac Sim, where we implement parallel training by collecting data from thousands of quadcopters simultaneously. The training environment features a forest-like setting with both static and dynamic obstacles. Each robot is spawned at a random location with a random goal and is reset either upon colliding with an obstacle or at the end of the episode. To improve learning efficiency, we adopt a curriculum learning strategy. The environment initially starts with a relatively low obstacle density, which is gradually increased as the success rate surpasses a specified threshold. Our experiments show that this approach allows the RL policy to achieve a higher navigation success rate in complicated environments.

\subsection{Policy Action Safety Shield} \label{safety shield}
\begin{figure}[t] 
    \centering
    \includegraphics[scale=0.60]{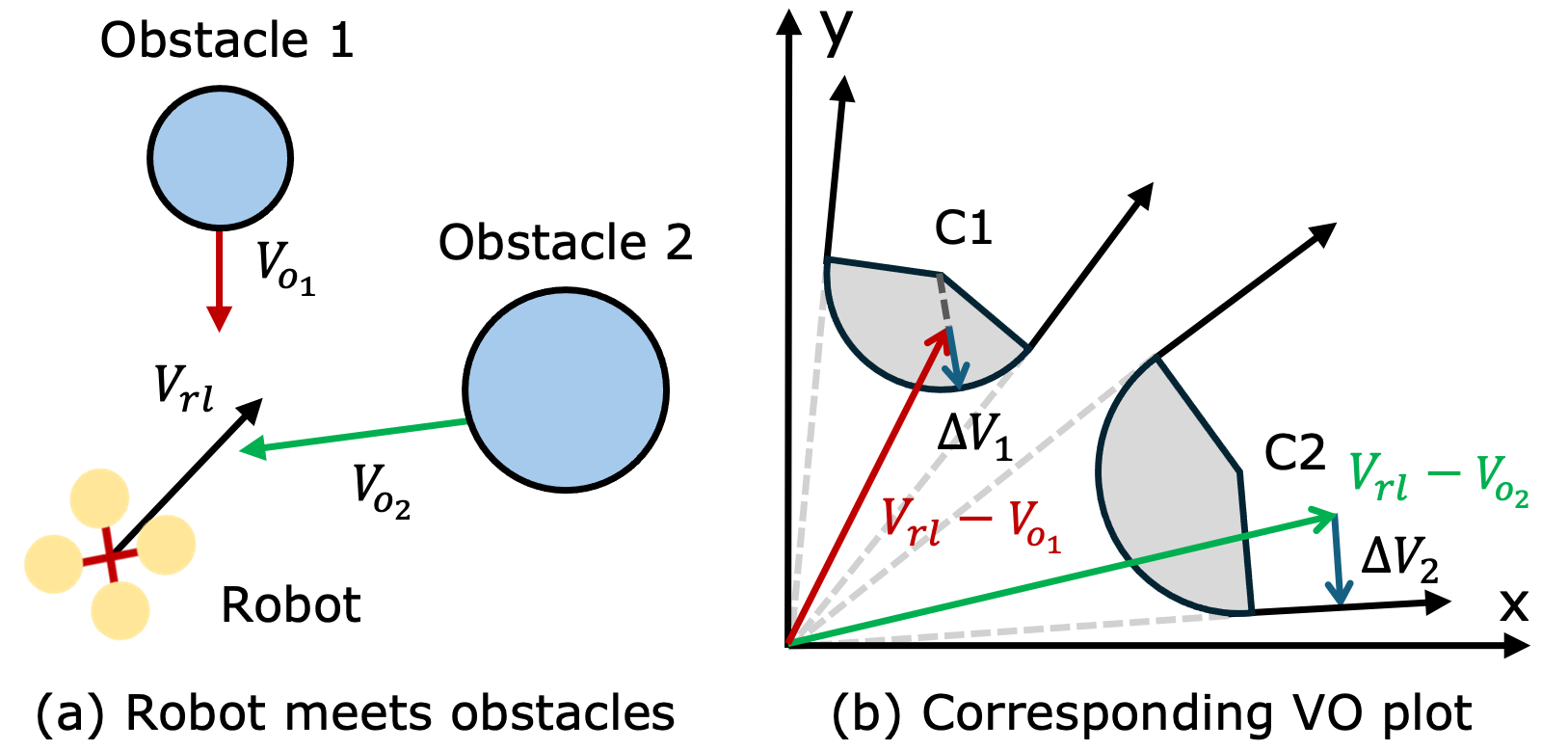}
    \caption{Illustration of determining the safe velocity region using the velocity obstacle-based method. (a) An example scenario where the robot encounters two obstacles. (b) The corresponding velocity obstacle plot with blue arrows showing the minimum velocity change required to exit the VO regions. }
    \label{VO-figure}
\end{figure}

Due to the black-box nature of the neural network, we designed a safety shield mechanism to prevent severe failures caused by the trained policy. Given that our policy outputs velocity commands for robot control, we incorporate the concept of velocity obstacles (VO) \cite{velocity_obstacles} to evaluate the safety of the policy's actions and correct them if they could result in collisions. The velocity obstacle represents the set of robot velocities that would result in a collision within a specified future time horizon. In the scenario depicted in Fig. \ref{VO-figure}a, where the robot encounters obstacles, the corresponding velocity obstacle regions are visualized in Fig. \ref{VO-figure}b. In this example, there are two obstacles moving at different velocities, each associated with a velocity obstacle region based on their relative positions and velocities with respect to the robot. Each velocity obstacle region combines a circular area centered at C1 or C2, with a radius equal to the sum of the robot and obstacle sizes plus user-defined safe space, and a cone area extending from the origin, excluding the dotted lines. The relative velocities of the robot to Obstacle 1 and Obstacle 2 (shown as red and green arrows) lie within the velocity obstacle regions, indicating future collisions. Inspired by \cite{orca}, we adopt a similar approach to compute the minimum changes in velocity ($\Delta V_{1}$ and $\Delta V_{2}$) required to exit velocity obstacle regions. The policy action
produced by the network is defined as the velocity $V_{rl}$. If this action is not within any of the velocity obstacle regions, we set the safe action $V_{safe}$ equal to the policy action. Otherwise, we formulate the following optimization to project the policy action into the safe region:

\begin{mini!}[2]
    {V_{safe} \in \mathbb{R}^3}{\begin{Vmatrix} V_{safe} - V_{rl} \end{Vmatrix},}{}{} 
\addConstraint{(V_{safe} - (V_{rl} - V_{o_{i}} + \Delta V_{i})) \cdot \Delta V_{i}}  \geq 0{}{} \label{plane constraint}
\addConstraint{V_{min} \leq V_{safe} \leq V_{max}}{} \label{control limits}
\addConstraint{\forall i \in \{1, \ldots , N\}},
\end{mini!}
where Eqn. \ref{plane constraint} defines a hyperplane based on the required changes in velocity to ensure the safe action lies outside the velocity obstacle regions, and Eqn. \ref{control limits} enforces the control limits on the safe action. The drawback of constraining the safe action to one side of the hyperplane can be overly conservative when many obstacles are present. However, since the policy action only fails occasionally, this conservatism does not significantly impact overall performance and, in most scenarios, helps ensure safety. For static obstacles, we use each cast ray to determine the obstacle's center position and radius, setting their velocity to zero. For dynamic obstacles, we enclose them using one or multiple spheres.

\section{Result and Discussion}
To evaluate the proposed framework, we present our training results under different configurations and conduct simulation and physical flight tests in various environments. The policy was trained in NVIDIA Isaac Sim on a NVIDIA GeForce RTX 4090 GPU for around 10 hours. The maximum velocity of the robot is set to 2.0 m/s. 
The simulation experiments are conducted on the RTX 4090 desktop, while computations for the physical flights are performed on our quadcopter's onboard computer (NVIDIA Jetson Orin NX). An Intel RealSense D435i camera is utilized for static and dynamic obstacle perception, and the LiDAR Inertial Odometry (LIO) algorithm \cite{9697912} is adopted for accurate robot state estimation. 
The static and dynamic feature extractors use 3-layer convolutional neural networks, outputting embeddings of sizes 128 and 64, respectively. The policy network consists of a two-layer multi-layer perceptron with a PPO clip ratio of 0.1. The ADAM optimizer is used with a learning rate of $5 \times 10^{-4}$. The reward discounting factor is set to 0.99.

\subsection{RL Training Results}
\begin{figure}[t] 
    \centering
    \includegraphics[scale=0.88]{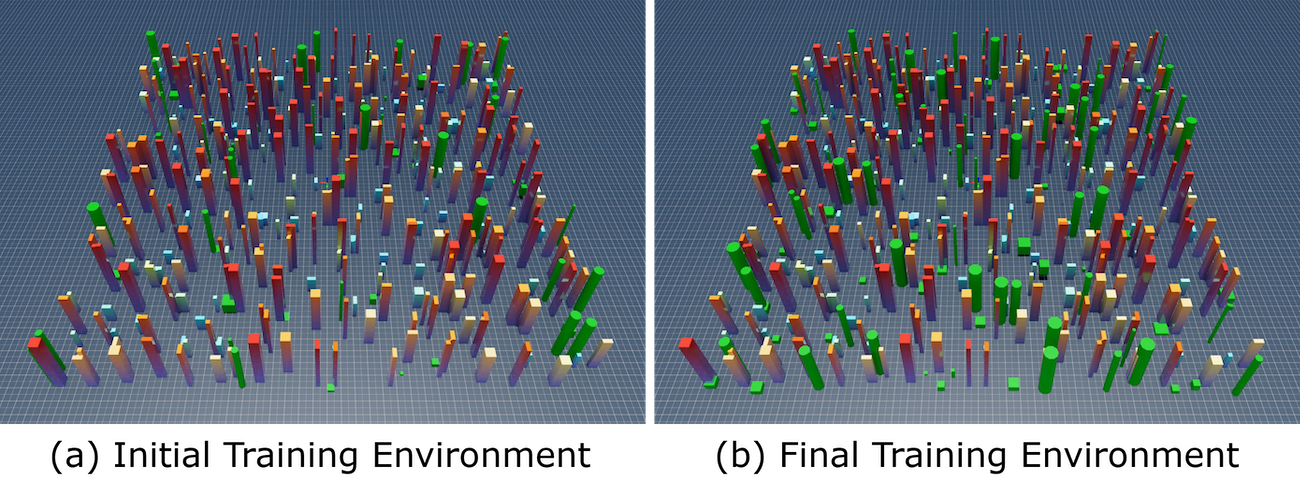}
    \caption{Visualization of robot training environments. The environment has a size of 50m$\times$50m, and the number of dynamic obstacles is gradually increased during training. (a) The initial environment contains 60 dynamic obstacles. (b) The final environment contains 120 dynamic obstacles. }
    \label{curriculum-learning}
\end{figure}

Our framework utilizes a curriculum learning strategy for our RL policy training, as illustrated by the initial and final training environments shown in Fig. \ref{curriculum-learning}. Static obstacles are shown in red with a color gradient indicating height, while dynamic obstacles are depicted in green. Based on our observations, dynamic obstacles present a greater challenge during training than static obstacles. Therefore, we gradually increase the number of dynamic obstacles in the environment from 60 to 120, in increments of 20, once the navigation success rate exceeds 80\% (defined as safely navigating from the start to the goal without collisions). To demonstrate the effectiveness of curriculum learning, Table \ref{curriculum-learning-table} compares the highest navigation success rates achieved with and without curriculum learning, keeping the total training time constant for both scenarios. The table shows that as the number of dynamic obstacles increases, the highest navigation success rate without curriculum learning decreases more sharply compared to training with curriculum learning, highlighting its effectiveness. We stopped policy training at 120 dynamic obstacles and saved the best model trained with 100 dynamic obstacles, as it achieved a relatively high success rate (80.96\%) under challenging conditions.

\begin{table}[t]
\begin{center}
\caption{Comparison of the highest navigation success rates during training with and without using curriculum learning.} \label{curriculum-learning-table}
\begin{tabular}{ |l | c | c | } 
 \hline

Number of Obstacles & No Curr. Learning & Curr. Learning \Tstrut\\ 
 \hline
 static=350, dynamic=60 & $94.33$\%   & $94.33$\%  \\   
 \hline
 static=350, dynamic=80 & $74.51$\%  & $82.71$\%\\   
 \hline  
 static=350, dynamic=100 & $62.30$\%  & $80.96$\% \\   
 \hline
 static=350, dynamic=120 & $54.98$\%  & $68.65$\% \\   
 \hline
\end{tabular}
\end{center}
\end{table}

\begin{figure}[t] 
    \centering
    \includegraphics[scale=0.115]{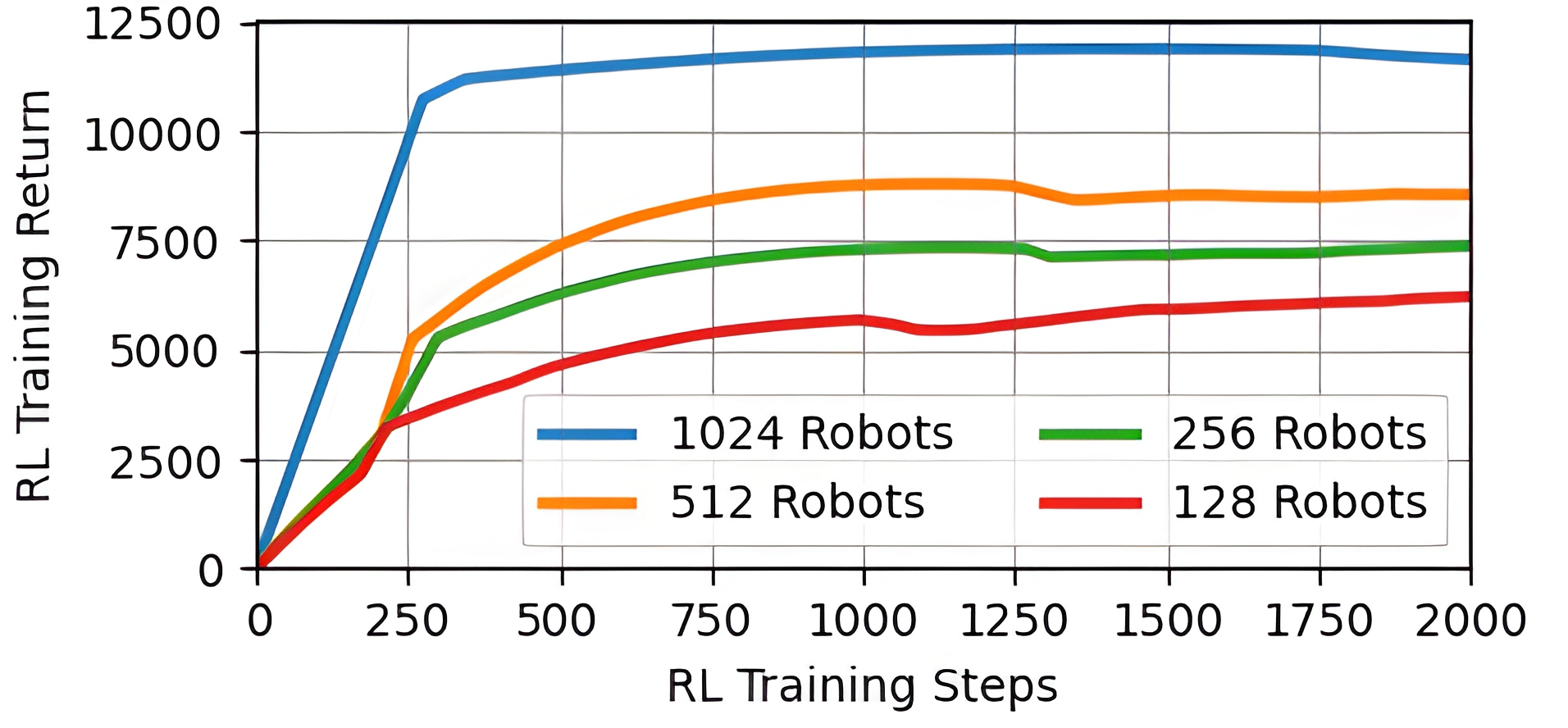}
    \caption{Plot of the smoothed average training return curves. Training with a larger number of robots leads to faster convergence and higher returns.}
    \label{training-curve}
\end{figure}

To show the importance of training with a larger number of robots, Fig. \ref{training-curve} compares the average RL training returns achieved with different numbers of robots deployed in training. The figure shows that training with more robots not only leads to faster convergence but also results in higher RL returns. Our experiments were conducted with 1024 robots, which utilized the maximum available GPU memory.

\begin{figure}[t] 
    \centering
    \includegraphics[scale=0.46]{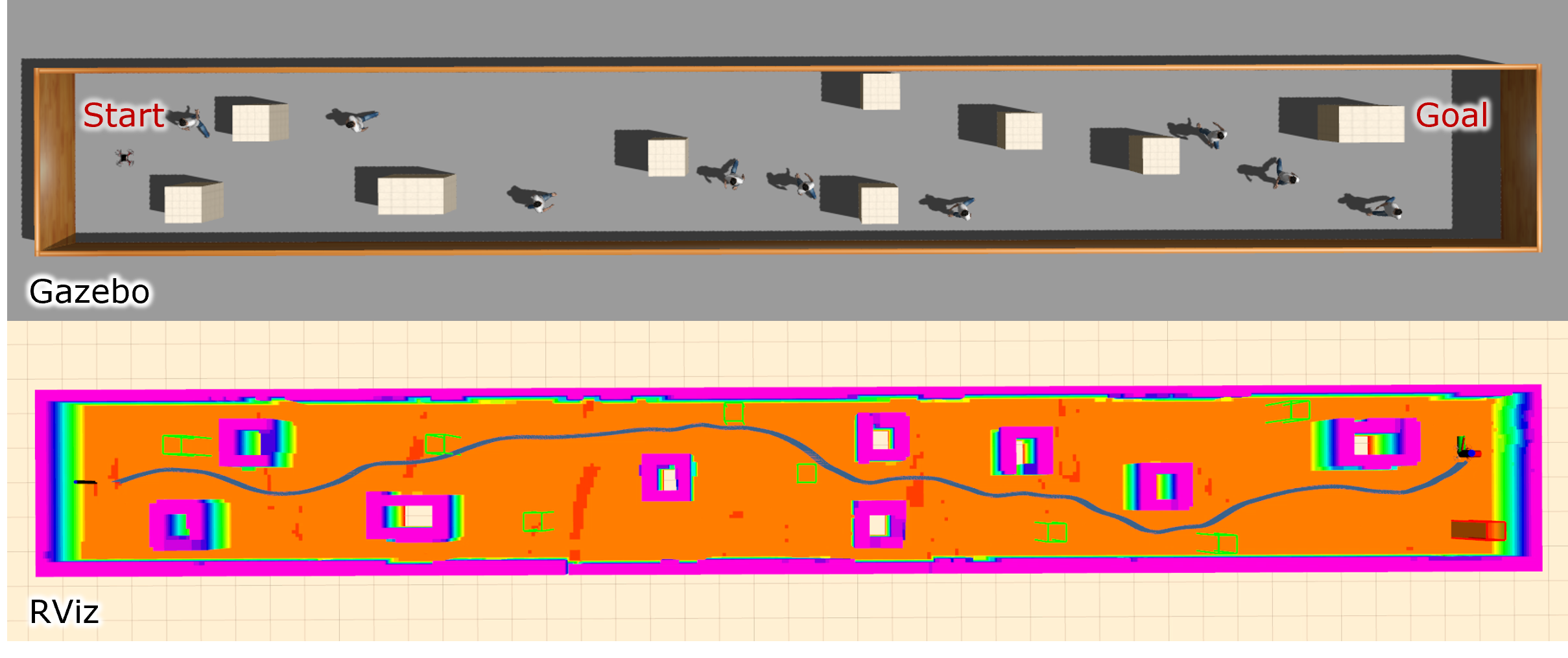}
    \caption{Visualization of a safe navigation trajectory in a simulated corridor environment. The top image illustrates the environment in Gazebo, containing static obstacles and dynamic obstacles (pedestrians). The bottom image displays the environment map with the robot's navigation trajectory. }
    \label{corridor-experiment}
\end{figure}

\subsection{Simulation Experiments}

We conducted simulation experiments in Gazebo to evaluate the proposed framework and demonstrate its sim-to-sim transferability. Quantitative tests were performed in indoor-like dynamic environments resembling real-world scenarios where the robot navigates safely around humans. Fig. \ref{corridor-experiment} shows an example of a safe trajectory. The results confirm that the robot successfully navigated without collisions.

To quantitatively evaluate the performance, we generate environments with high obstacle density, mirroring the conditions of our training environments, where both static and dynamic obstacles are randomly placed. Given the limited availability of open-source RL-based navigation benchmark algorithms, we compare our method with the popular optimization-based static planner \cite{ego_planner} and a vision-aided planner \cite{ViGO} designed for dynamic environments. Additionally, we evaluate the performance of our framework with and without the safety shield to verify its effectiveness. Specifically, we test in three types of environments: static, purely dynamic, and hybrid environments. Each algorithm is run 20 times in each environment, and the average number of collisions per run is calculated as shown in Table \ref{collision rate result}. The percentage value in Table \ref{collision rate result} indicates the average collision times relative to the baseline method \cite{ViGO}. It is important to note that these environments are designed to test the limits of the algorithms and are significantly more complex than real-world scenarios. Therefore, the occurrence of collisions does not necessarily indicate that the algorithm is unsafe. Overall, our NavRL demonstrates the lowest number of collisions in dynamic and hybrid environments, while maintaining a comparable collision rate to the EGO planner in static environments. The table shows N/A for the EGO planner in dynamic and hybrid environments because its inefficient map updates cause it to get stuck when handling excessive noise from dynamic obstacles. In the experiments, ViGO fails to provide sufficiently reactive trajectories for collision avoidance when obstacles are in close proximity to the robot. In contrast, our method can generate more reactive controls that help the robot avoid collisions more efficiently. Comparing our framework with and without the safety shield, we found the safety shield consistently reduced collisions, particularly in dynamic environments, where it mitigates errors caused by the neural network’s increased failure risk.



\begin{table}[h]
\begin{center}
\caption{Benchmark of the average number of collisions evaluated from 20 sample runs in different types of environments.} \label{collision rate result}
\begin{tabular}{ |l | c | c| c |  } 
 \hline

 \multicolumn{4}{|c|}{Average Collision Times Measurement in a 20m$\times$40m Map } \Tstrut\\
 \hline

 Benchmarks   & Static Env. & Dynamic Env. & Hybrid Env. \Tstrut\\
 \hline
 EGO \cite{ego_planner}  & 0.45 ($56.3\%$)  & N/A & N/A \Tstrut\\ 
 \hline
 ViGO \cite{ViGO} & 0.80 ($100\%$) & 3.15 ($100\%$) &  4.40 ($100\%$) \Tstrut\\  
 \hline
 Ours w/o Safe & 0.95 ($118.8\%$) & 2.70 ($85.7\%$) &  4.60 ($104.5\%$) \Tstrut\\  
 \hline
 \textbf{Ours (NavRL)}  & \textbf{0.65} ($\mathbf{81.3\%}$) & \textbf{0.85} ($\mathbf{27.0\%}$) & \textbf{2.10} ($\mathbf{47.8\%}$) \Tstrut\\ 
 \hline
\end{tabular}
\end{center}
\end{table}

\subsection{Physical Flight Tests}

\begin{figure}[t] 
    \centering
    \includegraphics[scale=0.57]{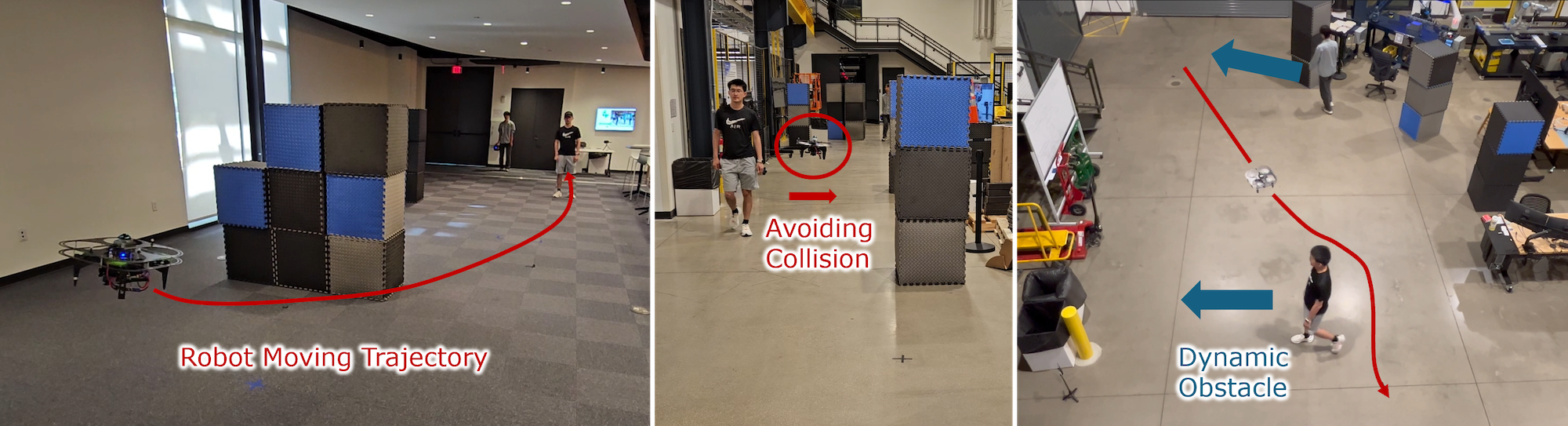}
    \caption{Examples of physical flight tests. Our framework enables the robot's safe flight and navigation in the presence of static and dynamic obstacles. }
    \label{real-experiment}
\end{figure}

To demonstrate sim-to-real transfer and safe navigation, we conducted physical flight experiments in various settings, as shown in Fig. \ref{real-experiment}. Static obstacles were placed in the environments, and several pedestrians were directed to walk toward the robot, requiring it to avoid collisions while reaching its goals. The results showed that the robot successfully avoided collisions and reached its destinations safely.

During the simulation and real-world flight experiments, we measured the computation time for each module in the proposed framework, as detailed in Table \ref{runtime table}. The table shows the runtimes on both the NVIDIA GeForce RTX 4090 and the onboard NVIDIA Jetson Orin NX computer. The static perception module completes in 8 ms on the RTX 4090 and 15 ms on the Orin NX, while the dynamic perception module takes 11 ms and 27 ms, respectively. The RL policy network runs in 1 ms on the RTX 4090 and 7 ms on the Orin NX, and the safety shield module operates in 2 ms and 16 ms. These measurements demonstrate that all modules are capable of real-time performance even on an onboard computer.

\begin{table}[t]
    \centering
    \caption{The runtime of each component of the proposed system.}
    \begin{tabular}{ l c c } 
    \hline
    System Modules & GeForce RTX 4090 & Jetson Orin NX\Tstrut\\
    \hline
    Static Perception &  8 ms & 15 ms \Tstrut\\ 
    Dynamic Perception & 11 ms & 27 ms \Tstrut\\ 
    RL Policy Network & 1 ms & 7 ms\Tstrut\\ 
    Safety Shield & 2 ms & 16 ms\Tstrut\\  
    \hline
    \end{tabular}
    \label{runtime table}
\end{table}

\section{Conclusion and Future Work}
This paper presents a novel deep reinforcement learning framework, NavRL, designed to achieve safe flight in dynamic environments based on the Proximal Policy Optimization (PPO) algorithm. The framework uses tailored state and action representations to enable safe navigation in both static and dynamic environments, supporting effective zero-shot sim-to-sim and sim-to-real transfer. Besides, A safety shield based on the velocity obstacle concept mitigates failures from the black-box nature of neural networks. Additionally, a parallel training pipeline with NVIDIA Isaac Sim accelerates the learning process. Results from simulation and physical experiments verify the effectiveness of our approach in achieving safe navigation within dynamic environments. Future work will focus on improving and adapting this framework for deployment across various robotic platforms.


\bibliographystyle{IEEEtran}
\bibliography{IEEEabrv,bibliography.bib}

\end{document}